\documentclass{article}

\usepackage{spconf,amsmath,graphicx}

\usepackage[T1]{fontenc} 
\usepackage[latin1]{inputenc}
\usepackage[frenchb,english]{babel}

\usepackage[capitalize,noabbrev]{cleveref} 
\crefname{equation}{equation}{equations}
\providecommand*{\fullref}[1]{\hyperref[{#1}]{\cref*{#1}. \nameref*{#1}}}
\providecommand*{\Fullref}[1]{\hyperref[{#1}]{\Cref*{#1}. \nameref*{#1}}}

\usepackage{amssymb} 
\usepackage{amsmath} 
\usepackage{amsfonts} 
\usepackage{mathrsfs}
\usepackage{amsthm}
\usepackage{kmath} 			
\usepackage{kfig} 			
\usepackage[noadjust]{cite}       
\usepackage{soul}		         

\usepackage{graphicx}
\usepackage{color}
\usepackage{pgfplots}
\usepackage{epsfig}
\usepackage{epstopdf}

\def\a{{\mathbf a}}

\def\c{{\mathbf c}}

\def\t{{\mathbf t}}
\def\u{{\mathbf u}}

\def\x{{\mathbf x}}
\def\y{{\mathbf y}}
\def\z{{\mathbf z}}
\def\0{{\mathbf 0}}
\def\1{{\mathbf 1}}

\def\A{{\mathbf A}}

\def\I{{\mathbf I}}

\def\zz{{\boldsymbol z}}

\def\Ac{{\mathcal A}}

\def\Gc{{\mathcal G}}

\def\Oc{{\mathcal O}}

\def\Rc{{\mathcal R}}

\def\Zc{{\mathcal Z}}

\def\Rbb{{\mathbb R}}

\def\thetab{{\boldsymbol \theta}}

\def\etal{\textit{et al.}\xspace }
\def\spa{\mathrm{span}}

\newkmacro\sign[1][{\cdot}]{\mathrm{sign}\left({#1}\right)} 
\newkmacro\interior[1][{\cdot}]{\mathrm{int}\left({#1}\right)}
\newkmacro\bd[1][{\cdot}]{\mathrm{bd}\left({#1}\right)}
\newkmacro\conv[1][{\cdot}]{\mathrm{conv}\left({#1}\right)}
\newkmacro\card[1][{\cdot}]{\mathrm{card}\kbrace{{#1}}}
\newkfunc{\projector}{P}

\newtheorem{lemma}{Lemma}

\def\rad{{\epsilon}}

\def\spc{{\alpha}}
\def\spcmin{{\delta}}
\def\obs{{\y}}
\def\dimobs{{m}}
\def\sv{{\x}}
\def\optimvar{{\u}}
\def\dimx{{n}}
\def\dico{{\Ac}}
\def\dicomat{{\A}}
\def\atom{{\a}}
\def\cen{{\c}}
\def\Hs{{\Rbb}^\dimobs}

\def\radss{{\tau}}
\def\tat{{\t}}
\def\tatt{{\tilde{\tat}}}
\def\rdx{{l}}
\def\rmax{{L}}
\def\sreg{{\Rc}}
\def\treg{\Gc}
\def\tregs{\treg^s}
\def\tregd{\treg^d}
\def\targetset{{\cal{S}}}

\newkfunc{\support}{\mathrm{supp}}
\newkfunc{\acts}{\mathrm{acts}}
\newkmacro{\scap}[2][]{\kangle{#1,#2}}
\newkmacro{\nor}[1][]{\kvvbar{#1}_2}

\title{JOINT SCREENING TESTS FOR LASSO}

\twoauthors
  {C\'edric Herzet$^{*}$}
	{INRIA Rennes-Bretagne Atlantique\\
	Campus de Beaulieu\\
	 35000 Rennes, France}
  {Ang\'elique Dr\'emeau\sthanks{CH was supported by the French National Research Agency through the GERONIMO and BECOSE projects. AD was supported by the French Defense Procurement Agency (DGA) through the TS-DECO MRIS project. CH is currently working in Lab-STICC (UMR 6285), France.}}
	{ENSTA Bretagne \& Lab-STICC \\
	2 rue Francois Verny\\
	29200 Brest, France}

\begin{document}
%
\maketitle
\begin{abstract}
This paper focusses on ``safe'' screening techniques for the LASSO problem. Motivated by the need for low-complexity algorithms, we propose a new approach, dubbed ``joint screening test'', allowing to screen a set of atoms by carrying out one single test. The approach is particularized to two different sets of atoms, respectively expressed as sphere and dome regions. After presenting the mathematical derivations of the tests, we elaborate on their relative effectiveness and discuss the practical use of such procedures.
\end{abstract}

\begin{keywords}
$\ell_1$-norm minimization, LASSO, screening techniques.
\end{keywords}

\section{Introduction}

In the last decade, sparse representations have proven to be powerful tools for solving many problems in signal processing, machine learning, etc.~Many central methodologies to find a good sparse representation of an observation vector $\obs\in\Hs$ in some dictionary $\dicomat = \kbracket{\atom_1 \ldots \atom_\dimx}\in\Rbb^{\dimobs\times\dimx}$ revolve around the resolution of the so-called 
(nonnegative) LASSO problem:\footnote{The standard LASSO can be written as a particular case of \eqref{eq:Primal_Problem}, see \cite{Figueiredo2014Teaching}.}
\begin{align}\label{eq:Primal_Problem}
\x^\star_\lambda \in \kargmin_{\x \geq \0}\, P_\lambda(\obs,\x),
\end{align}
where $P_\lambda(\obs,\sv)= \frac{1}{2}\nor[{\obs- \dicomat \x }]^2+\lambda \kvvbar{\x}_1$ and $\x\in\Rbb^\dimx$. Without loss of generality, we will assume hereafter that $\nor[\atom_i]=1$ for $i=1\ldots \dimx$. 

Solving \eqref{eq:Primal_Problem} may require a heavy computational load when the dimension of $\sv$ becomes large. 
 Therefore, the conception of computationally-efficient techniques to solve \eqref{eq:Primal_Problem} has become an active field of research \cite{Osborne2000New,Efron2004Least,Beck2009Fast}. One important contribution in this field is the so-called ``safe'' \footnote{The term ``safe'' refers to the fact that the elements identified by the screening method always correspond to zeros in $\x^\star_\lambda$.} screening technique proposed by El Ghaoui \etal in \cite{EVR:10} and refined in several subsequent works \cite{NIPS2011_0578,Dai2012Ellipsoid,wang2015lasso,Bonnefoy2015Dynamic,Fercoq_Gramfort_Salmon15,Herzet16Screening,Xiang2012Fast,Xiang2014Screening}. 
 These methodologies aim at decreasing the dimensionality of the problem to handle by identifying some of the zeros of the target solution $\x^\star_\lambda$ via simple tests. 
 Screening procedures have been shown to allow for a dramatic reduction of the complexity needed to solve~\eqref{eq:Primal_Problem}.  Nevertheless, their implementation may still be computationally too-demanding in some applications. More specifically, the complexity of all the screening procedures proposed so far evolves linearly with the number of atoms in the dictionary. 
Hence, in applications involving very large dictionaries,\footnote{As an extreme example, one may think of the ``continuous'' dictionaries considered in \cite{Candes2014Towards,Duval2015}, containing an infinite number of atoms.} the implementation of these screening tests may lead to unacceptable computational burdens. 
%
 %
 In this paper, we propose a new screening methodology, dubbed  ``joint screening'', allowing to circumvent this issue.  
\section{Some Convex Considerations }\label{sec:stdscreening}

We first remind some of the properties of the solutions of \eqref{eq:Primal_Problem}. 
%
Problem \eqref{eq:Primal_Problem} is convex and always admits (at least) one solution. The dual problem associated to  \eqref{eq:Primal_Problem} can be written as (see for example \cite{Foucart2013Mathematical})
 \begin{align}\label{eq:Dual_Problem}
 \thetab^\star_\lambda = \kargmin_{\thetab\in\mathcal{D}} D_\lambda(\obs,\thetab),
 \end{align}
 where 
 \begin{align}
 D_\lambda(\obs,\thetab) &=  \frac{1}{2}\nor[{\obs}]^2-\frac{1}{2}\nor[{\obs-\lambda\thetab}]^2,\\
 \mathcal{D} &= \{\thetab\in\Hs : \scap[\atom_i]{\thetab} \leq 1,\, i = 1 \ldots n\}, \label{eq:dual_set}
 \end{align}
 and $\scap[\cdot]{\cdot}$ denotes the inner product in $\Rbb^\dimobs$. 
 Since $D_\lambda(\obs,\thetab)$ is a strictly concave and coercive function and $\mathcal{D}$ is a closed set, problem \eqref{eq:Dual_Problem} admits a unique solution $\thetab^\star_\lambda$, see \cite[Proposition A.8]{Bertsekas_book}.

The primal and dual solutions $(\x^\star_\lambda,\thetab^\star_\lambda)$ are related through the Karush-Kuhn-Tucker conditions \cite[Proposition 5.1.5]{Bertsekas_book}:
\begin{align}
&\x^\star_\lambda\geq \0,\  \scap[\a_i] {\thetab^\star_\lambda} \leq 1 \mbox{ for all $i$},\label{eq:primal feasibility}\\ 
&\kparen{\scap[\a_i] {\thetab^\star_\lambda}-1} \x^\star_\lambda (i) =0\ \mbox{ for all $i$}, \label{eq:complementary slackness} \\
&\y = \lambda \thetab^\star_\lambda + \A \x^\star_\lambda, \label{eq:primal feasibility 2} 
\end{align}
where 
$\x^\star_\lambda (i)$ denotes the $i$th component of  $\x^\star_\lambda$.

\section{Standard Screening Methodologies}\label{sec:stdscreening}

The ``safe'' screening procedures proposed in \cite{EVR:10,NIPS2011_0578,Dai2012Ellipsoid,wang2015lasso,Bonnefoy2015Dynamic,Fercoq_Gramfort_Salmon15,Herzet16Screening,Xiang2012Fast,Xiang2014Screening} leverage on the following observation: if $\sreg\subset\Rbb^\dimobs$ is a region such that $\thetab^\star_\lambda\in \sreg$,\footnote{Such a region is commonly referred to as ``safe region'' in the screening literature.} then the following inequality trivially holds 
\begin{align}
\scap[\a_i] {\thetab^\star_\lambda} \leq \max_{\thetab\in\sreg} \scap[\atom_i]{\thetab}, \nonumber
\end{align}
and from \eqref{eq:complementary slackness}, we thus have
\begin{align}
\max_{\thetab\in\sreg} \scap[\atom_i]{\thetab} < 1 \Rightarrow \x_\lambda^\star(i)=0. \label{eq:gen_screening_test}
\end{align}
%
%
In other words, if the inequality in the left-hand side of \eqref{eq:gen_screening_test} is satisfied, one is ensured that the $i$th component of the solution vector $\x_\lambda^\star$ is equal to zero. 

Since the seminal work by El Ghaoui \etal, different screening tests, based on different choices of  the region $\sreg$, have been proposed in the literature. The most popular ones are probably the ``sphere'' regions, that is 
\begin{align}\label{eq:safe_sphere}
\sreg&=\kbrace{\thetab: \nor[\thetab-\cen]\leq 1-\radss}, 
\end{align}
for some parameters $\cen\in\Hs$ and $\radss\leq1$. Interestingly, for this particular choice, the general screening test \eqref{eq:gen_screening_test} takes the following simple form:
 \begin{align}
 \scap[\atom_i]{\cen}< \radss \Rightarrow \x_\lambda^\star(i)=0. \label{eq:test} 
\end{align} 
We find in the literature different definitions for the center $\cen$ and the radius $1-\radss$ (see \cite{EVR:10, NIPS2011_0578,Dai2012Ellipsoid,wang2015lasso,Bonnefoy2015Dynamic,Fercoq_Gramfort_Salmon15,Herzet16Screening}) leading to screening tests of different effectiveness. 

It is easy to see that the implementation of the standard screening test \eqref{eq:gen_screening_test} necessitates the evaluation of $\max_{\thetab\in\sreg} \scap[\atom_i]{\thetab}$ for \textit{each} atom of the dictionary.~Hence, the computational load required to implement standard screening tests evolves linearly with the number of atoms in the dictionary. For example, in the case where $\sreg$ is a ``sphere'' region, \eqref{eq:test} involves the computation of the inner product between the center of the sphere $\cen$ and each atom of the dictionary,  leading to a complexity scaling as $\mathcal{O}(\dimobs\dimx)$. 
 In the next section, we propose a new procedure to reduce this computational load. 

%


\section{Joint Screening Procedures}\label{sec:groupscreening}

In this section, we introduce a new screening procedure having a complexity not depending on the number of atoms in the dictionary. We dub our methodology ``joint screening test'' because it allows to screen a \textit{set} of atoms by carrying out one \textit{single} test. In a first subsection, we derive tests allowing to screen any atom belonging to some specific region $\treg\subset\Rbb^\dimobs$. In a second subsection, we elaborate on the relative effectiveness of the proposed test for different choices of the region $\treg$.  Finally, in the last subsection, we leverage on these previous results to propose a novel screening procedure having a low complexity with regard to standard screening tests.


\subsection{Screening the atoms belonging to a region $\treg$}

\begin{figure}
\begin{kfig}[0.8\columnwidth]{1}
\includegraphics{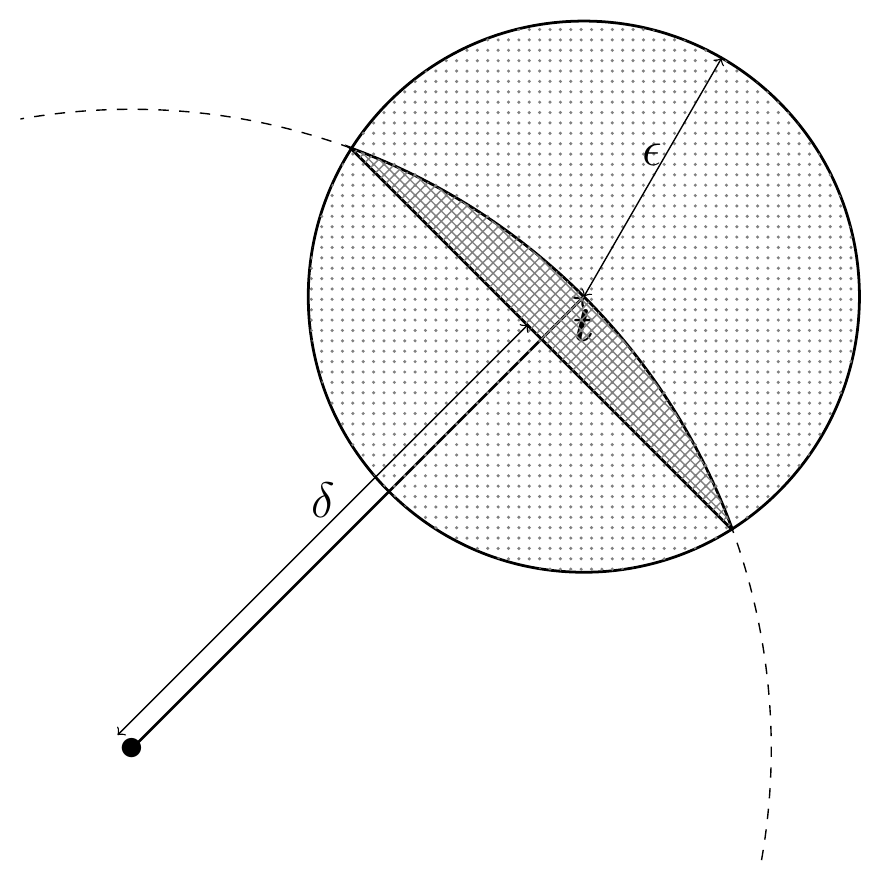}
\end{kfig}
\caption{\small Illustration of the sphere region $\tregs(\tat,\rad)$ (dotted area) and the dome region $\tregd(\tat,\spcmin)$ (cross-hatched area). The region inside the dashed curves corresponds to set of atoms with a $\nor[\cdot]$-norm smaller than 1. }\label{fig:geomrepr}
\end{figure}

Let $\dico = \kbrace{\atom_i}_{i=1}^{\dimx}$ denote the set of atoms of the dictionary and  let $\sreg\subset\Hs$ be a safe region (that is  $\thetab^\star_\lambda\in\sreg$). 
The ``joint'' screening procedure proposed in this paper is a direct consequence of the following observation:\footnote{The validity of \eqref{eq:group_gen_screening_test} straightforwardly follows from \eqref{eq:gen_screening_test}.} 
\begin{align}
\max_{\atom\in\treg}\max_{\thetab\in\sreg} \scap[\atom]{\thetab} < 1 \Rightarrow \x_\lambda^\star(i)=0\quad \forall i : \atom_i\in\dico\cap\treg. \label{eq:group_gen_screening_test}
\end{align}
In order words, if the inequality in the left-hand side of \eqref{eq:group_gen_screening_test} is satisfied, all the atoms $\atom_i\in\dico\cap\treg$ can be \textit{safely} and \textit{jointly} screened from problem \eqref{eq:Primal_Problem}.  

In what follows, we will see that the verification of the inequality in the left-hand side of \eqref{eq:group_gen_screening_test} can be done very efficiently for some specific choices of regions $\sreg$ and $\treg$. First, we will assume that $\sreg$ is a sphere region \eqref{eq:safe_sphere}.
The joint screening test \eqref{eq:group_gen_screening_test} then takes the simple form:
\begin{align}
\max_{\atom\in\treg}  \scap[\atom]{\cen} < \radss \Rightarrow \x_\lambda^\star(i)=0\quad \forall i :  \atom_i\in\dico\cap\treg. \label{eq:group_gen_screening_test_sphere}
\end{align}
%
Moreover, we will consider the two following options for $\treg$:
\begin{itemize}
\item \textit{``Sphere'' }: $\tregs(\tat,\rad) = \kbrace{\atom : \nor[\atom - \tat]\leq \rad}$,
\item \textit{``Dome'' }\ : $\tregd(\tat,\spcmin) = \kbrace{\atom : \scap[\atom]{\tat} \geq \spcmin, \nor[\atom]\leq 1}$,
\end{itemize}
where $\tat\in\Hs$ 
and  $\rad$, $\spcmin$ are some parameters. $\tregs(\tat,\rad)$ and $\tregd(\tat,\spcmin)$ have some easy geometric interpretations: $\tregs(\tat,\rad)$ corresponds to the set of vectors located in a ball of radius $\rad$ centered on $\tat$;   $\tregd(\tat,\spcmin)$ is a dome including all the  vectors of norm smaller than one and having an inner product with $\tat$ greater than or equal to $\spcmin$. A graphical representation is given in Fig.~\ref{fig:geomrepr}.

For these two choices of regions, the joint screening test defined in \eqref{eq:group_gen_screening_test_sphere} admits the following simple analytical solutions:
\begin{itemize}
\item \textit{Joint sphere test: } $\displaystyle{\max_{\atom\in\tregs(\tat,\rad)} \scap[\atom]{\cen} < \tau}$ if and only if 
\begin{align}
\scap[\tat]{\cen} <  \radss -\rad\, \nor[\cen]. \label{eq:spheretest}
\end{align}
\item \textit{Joint dome test: } $\displaystyle{\max_{\atom\in\tregd(\tat,\spcmin)} \scap[\atom]{\cen} < \tau}$ if and only if \footnote{We assume that $\nor[\tat]=1$.} 
\begin{align}
\scap[\tat]{\cen} < \radss, \label{eq:dometest1}
\end{align}
and 
\begin{align}
\spcmin > \frac{\scap[\tat]{\cen} \radss + \sqrt{\nor[\cen]^2-\scap[\tat]{\cen}^2} \sqrt{\nor[\cen]^2-\radss^2}}{\nor[\cen]^2}.
\label{eq:dometest2}
\end{align}
\end{itemize}
A proof of this result can be found in Appendix~\ref{app:defsup}.

\subsection{Relative effectiveness of the screening tests}\label{sec:Relative effectiveness}

It is easy to see from \eqref{eq:group_gen_screening_test} that the choice of region $\treg$ is a compromise between the number of atoms that can be jointly screened and the ease of passing the test. Indeed, although large regions allow to screen more atoms, they are also less likely to pass the joint screening test since, for any $\treg_1\subseteq\treg_2$, we have
\begin{align} \label{eq:releffectiveness}
\max_{\atom\in\treg_1}\max_{\thetab\in\sreg} \scap[\atom]{\thetab}\leq \max_{\atom\in\treg_2}\max_{\thetab\in\sreg} \scap[\atom]{\thetab}.
\end{align}
In particular, letting $\treg_1=\kbrace{\atom_i}$ and $\treg_2=\treg$ in the above inequality, we see that passing the joint screening test \eqref{eq:group_gen_screening_test} requires in particular that the standard screening test \eqref{eq:gen_screening_test} is verified for any atom $\atom_i\in\dico\cap\treg$. Hence, quite logically, joint screening test \eqref{eq:group_gen_screening_test} can only lead to inferior screening performance as compared to standard screening test \eqref{eq:gen_screening_test}. 


 Another question of interest is the relative effectiveness of the joint sphere and dome tests proposed in \eqref{eq:spheretest} and \eqref{eq:dometest1}-\eqref{eq:dometest2}, respectively. The next lemma provides some insights into this question.
\begin{lemma}\label{lemma:inclusion_sets}
The smallest\footnote{``Smallest'' should  be understood as the one with the smallest volume.} dome containing a set of unit-norm vectors $\targetset$ is always contained in the smallest sphere containing $\targetset$.
\end{lemma}
A proof of this statement can be found in Appendix~\ref{app:proof_relation_mindome_minsphere}.~In view of \eqref{eq:releffectiveness}, a direct consequence of Lemma \ref{lemma:inclusion_sets} is as follows:  if one wishes to jointly screen a set of unit-norm atoms, there always exists a joint dome test leading to screening performance at least as good as the ``best''\footnote{``Best'' should be understood as the test involving the sphere of smallest volume.}  joint sphere test.

\subsection{Low-complexity screening procedures}\label{sec:loxcomplexity_st}

In this section, we discuss the following screening procedure:
\begin{itemize}
\item[1)] Select a set of $\rmax$ (dome or sphere) regions $\kbrace{\treg_{\rdx}}_{\rdx=1}^\rmax$, 
\item[2)] Apply screening test \eqref{eq:spheretest} or \eqref{eq:dometest1}-\eqref{eq:dometest2} for $\rdx = 1\ldots\rmax$. 
\end{itemize}
We note that since \eqref{eq:spheretest} and \eqref{eq:dometest1}-\eqref{eq:dometest2} only involve the evaluation of one inner product (namely $\scap[\tat]{\cen}$), the  screening procedure described above only requires to carry out a total number of $\rmax$ inner products. 

 On the other hand, the effectiveness of this screening procedure obviously depends on the choice of the regions $\kbrace{\treg_{\rdx}}_{\rdx=1}^\rmax$. 
 In order to find a trade-off between the two conflicting objectives emphasized in the first point of the previous section, we consider hereafter some specific choices of the regions $\kbrace{\treg_{\rdx}}_{\rdx=1}^\rmax$.  

Let us focus on one particular region $\treg_\rdx$ for a given $\rdx$. We first assume that $\treg_{\rdx}$ is a sphere, that is $\treg_{\rdx}=\tregs(\tat,\rad)$. Given the value of $\tat$, we want to tune the value of $\rad$ so that $\treg$ is the largest sphere passing (if possible) the joint screening test \eqref{eq:spheretest}. We note then that
the joint sphere test \eqref{eq:spheretest} is satisfied as soon as the radius $\rad$ verifies
\begin{align}
\rad<\rad_{\tat,\cen} \triangleq \frac{\radss-\scap[\tat]{\cen}}{\nor[\cen]} . \label{eq:optradius}
\end{align}
Now, letting the radius $\rad$ of $\treg_\rdx$ tend to its largest value $\rad_{\tat,\cen}$, we have that any atom $\atom_i\in \Ac$ having a distance to $\tat$ strictly smaller than $\rad_{\tat,\cen}$ will be screened by test \eqref{eq:spheretest}, that is 
\begin{align}
\nor[\atom_i-\tat] <\rad_{\tat,\cen} \Rightarrow \x_\lambda^\star(i)=0. \label{eq:jointscreeningtestsphere}
\end{align}
We note that \eqref{eq:jointscreeningtestsphere} defines a screening test on \textit{all} the elements of the dictionary (it may be applied to any atom $\atom_i\in\dico$) although it only requires the evaluation of one inner product $\scap[\tat]{\cen}$ (to evaluate $\rad_{\tat,\cen}$).~Let us moreover mention that the quantities  $\kbrace{\nor[\atom_i-\tat]}_{i=1}^\dimx$ 
 can be precomputed and sorted 
 once for all in advance, so that the identification of the atoms verifying \eqref{eq:jointscreeningtestsphere} can be done very efficiently (one may for example achieve a complexity scaling as $\Oc(\log_2 \dimx)$). As a consequence, the ``on-line'' complexity associated to the implementation of \eqref{eq:jointscreeningtestsphere} is of the order of $\Oc(\dimobs+\log_2 \dimx)$. 

 
 \begin{figure*}
\centering
\begin{kfig}[0.99\linewidth]{2}
\includegraphics[scale = 1.5]{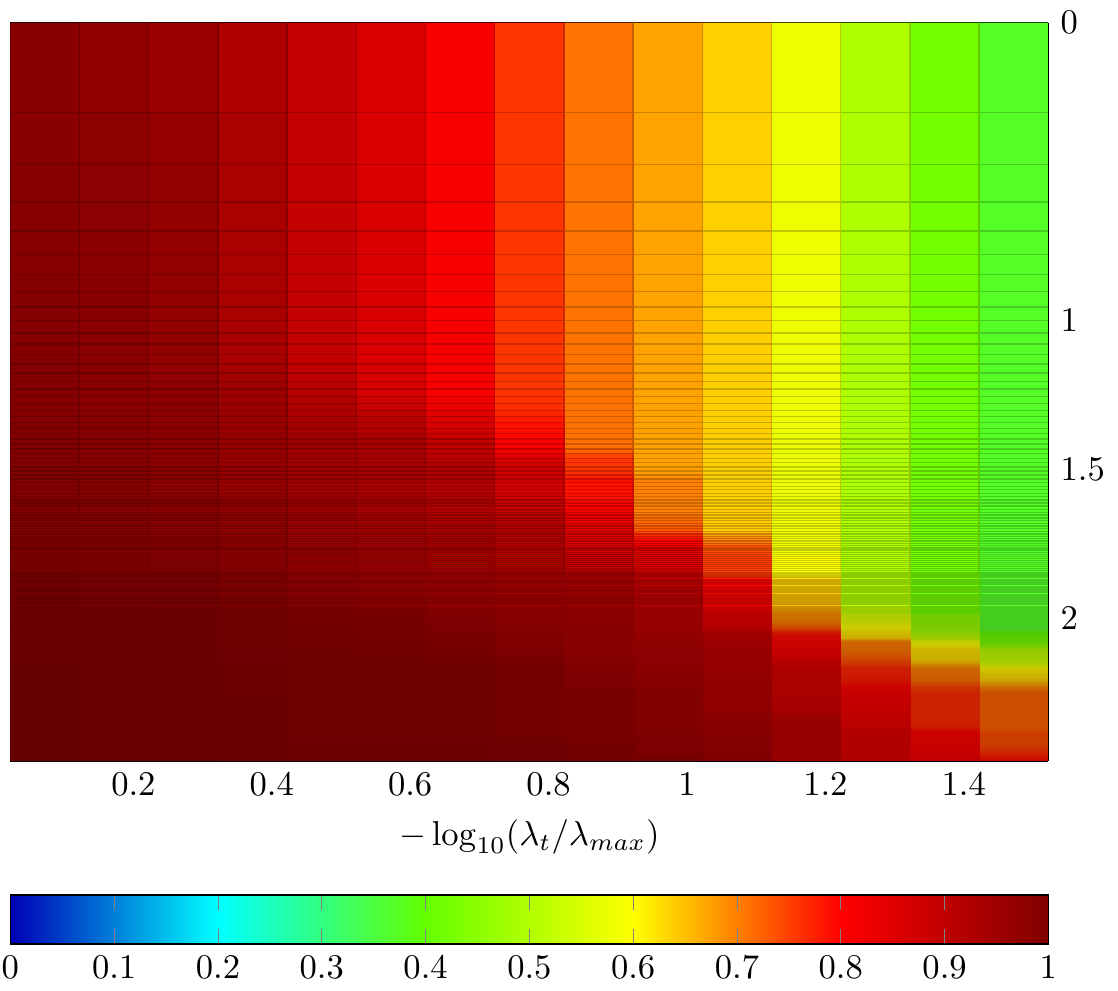}& \includegraphics{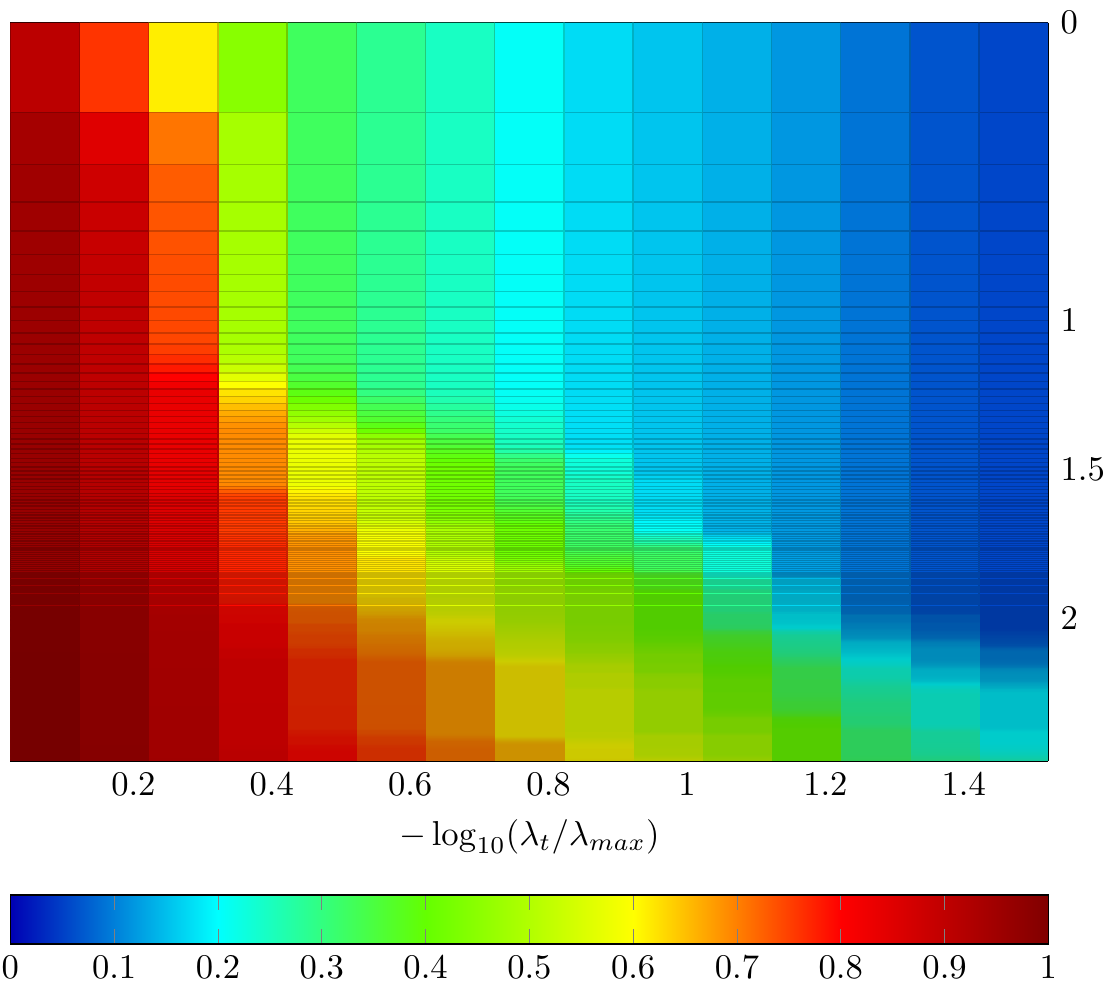}  
\end{kfig}
\caption{Proportion of zeros identified by the screening procedures as a function of $-\log_{10}(\lambda_t/\lambda_{max})$ (horizontal axis) and the ($\log_{10}$ of the) number of iterations (vertical axis): GAP sphere test (left) and proposed procedure with a dome region (right) \label{fig:results}}
\end{figure*}

 
We can apply the same kind of reasoning when $\treg_\rdx$ is a dome region, that is $\treg_\rdx=\tregd(\tat,\spcmin)$. If we assume that $\tat\in\Rbb^{\dimobs}$ is given, a tight lower bound on the value of $\spcmin$ verifying the joint dome test is trivially given by the right-hand side of \eqref{eq:dometest2}, that is 
\begin{align}
\spcmin>\spcmin_{\tat,\cen},
\end{align}
where
\begin{align}
\spcmin_{\tat,\cen} \triangleq \frac{\scap[\tat]{\cen} \radss + \sqrt{\nor[\cen]^2-\scap[\tat]{\cen}^2} \sqrt{\nor[\cen]^2-\radss^2}}{\nor[\cen]^2}.
\end{align}
 Hence, provided that $\scap[\tat]{\cen} < \radss$, letting parameter $\spcmin$ tend to its smallest value $\spcmin_{\tat,\cen}$ will lead to the  screening of any atom $\atom_i\in\dico$ having an inner product with $\tat$ strictly greater than $\spcmin_{\tat,\cen}$,
  that is:
\begin{align}
\left\{
\begin{array}{l}
\scap[\tat]{\cen} < \radss\\
\scap[\tat]{\atom_i} > \spcmin_{\tat,\cen}
\end{array}
\right.
\Rightarrow \x_\lambda^\star(i)=0. \label{eq:jointscreeningtestdome}
\end{align}
Again, the quantities $\kbrace{\scap[\tat]{\atom_i}}_{i=1}^\dimx$ 
can be precomputed and sorted once for all in advance, so that a complexity scaling as $\Oc(\dimobs+\log_2 \dimx)$ can also be achieved here. 

Going back to the screening procedure advocated at the beginning of this section, we see that adapting the parameter $\rad$ (resp. $\spcmin$) for each region $\treg_{\rdx}$ as discussed above is equivalent to applying test \eqref{eq:jointscreeningtestsphere} (resp. \eqref{eq:jointscreeningtestdome}) for each of the $\rmax$ different test vectors $\tat$ specifying the regions $\kbrace{\treg_{\rdx}}_{\rdx=1}^\rmax$. The overall complexity of this procedure thus scales as $\Oc( \rmax\dimobs+\rmax\log_2 \dimx)$. This is to compare to the complexity in $\Oc(\dimobs\dimx)$ of the standard screening tests. The joint screening procedures proposed in this paper will then be of particular interest when dealing with high-dimensional dictionaries.

\section{Numerical experiments}\label{sec:simus}

In this section we perform some numerical experiments to evaluate the behavior of the proposed method. We confront our screening procedure to the standard one  \eqref{eq:gen_screening_test} within the following setup.~We consider a dictionary $\dicomat\in\Rbb^{100\times2000}$ made up of $\rmax = 100$ clusters of $20$ atoms. For each cluster, a ``seed'' atom is created as the $\dimobs$-dimensional realization of a zero-mean circular Gaussian with covariance matrix $\dimobs^{-1} \I_\dimobs$, where $\I_\dimobs$ is the $\dimobs\times\dimobs$ identity matrix. The other atoms of the cluster are generated so that their inner products with the seed atom are not smaller than $0.9$. 
We assume that $\obs$ is a linear combination of $10$ columns (chosen randomly) of the dictionary. The nonzero coefficients are generated as independent realizations of a zero-mean Gaussian with variance equal to~$1$.

Considering this data set, we target the solution of problem \eqref{eq:Primal_Problem} for a decreasing sequence of $\lambda$ going from $\lambda_{\mathrm{max}}\triangleq \kvvbar{\ktranspose{\dicomat}\obs}_\infty$ to $10^{-1.5}\lambda_{\mathrm{max}}$.  The solution is searched via the FISTA algorithm~\cite{Beck2009Fast}. 
At each iteration of FISTA, the screening procedure is implemented as follows.~A ``safe'' sphere $\sreg$ is computed according to the GAP procedure proposed in \cite{Fercoq_Gramfort_Salmon15}. $\sreg$ is then used to implement both a standard screening test \eqref{eq:gen_screening_test} and the reduced-complexity screening test presented in Section \ref{sec:loxcomplexity_st}. Due to space limitation, we only consider the results obtained with the test based on the dome region,\footnote{The results corresponding to the test based on the sphere region \eqref{eq:jointscreeningtestsphere} are however sensibly similar.} see~\eqref{eq:jointscreeningtestdome}. The $\rmax$ test vectors appearing in \eqref{eq:jointscreeningtestdome} are set to be equal to the ``seed'' vectors used to generate each cluster of the dictionary. 

We evaluate the performance of the methods as the good detection rate of zeros in the solution vector $\x^\star_\lambda$.~Fig.~\ref{fig:results} presents this figure of merit as a function of $\lambda$ (horizontal axis) and the iteration number (vertical axis). 
As expected (see Section \ref{sec:Relative effectiveness}), the standard screening test (left figure) presents better performance than the proposed methodology (right figure).~However, this performance must be weighed against the complexity required to perform the tests: the standard procedure requires 2000 scalar products for each test, whereas the proposed method involves only 100 scalar products. We thus see that the proposed procedure allows for a good compromise between computational complexity and the ability to identify zeros of $\x^\star_\lambda$. 

\section{Conclusions}

In this paper, we proposed a new screening methodology for the (nonnegative) LASSO problem.~Our procedure aims to jointly screen a set of similar atoms by carrying out one single test.~We considered two instances of such test (focussing on sets of atoms belonging to either a sphere or a dome region of $\Rbb^\dimobs$) and showed that the latter take a very simple form. In particular, both tests only require the evaluation of \textit{one} inner product in $\Rbb^{\dimobs}$. Leveraging on this result, we showed that screening procedures for the entire dictionary can be devised by considering an arbitrary number, say $\rmax$, of regions. The resulting screening procedure has a complexity scaling as $\Oc(\rmax\dimobs+ \rmax\log_2\dimx)$, where $\dimx$ is the number of atoms in the dictionary. This has to be compared to the complexity of standard screening procedures scaling as $\Oc(\dimobs\dimx)$. 

Our future avenues of research include designing new joint tests (based on more refined regions $\treg$) and devising low-complexity methodologies to identify safe regions $\sreg$.

\appendix
\section{Derivation of the joint screening tests} \label{app:defsup}

 In this appendix, we provide a the detailed derivations leading to the joint screening tests stated in  \eqref{eq:spheretest} and \eqref{eq:dometest1}-\eqref{eq:dometest2}. Some steps of our reasoning are based on the technical lemmas stated in Appendix~\ref{app:technical_lemmas}.

\subsection{Sphere joint test \eqref{eq:spheretest}}

Let us first notice that the sphere region $\tregs(\tat,\rad)$ can be written as
\begin{align}
\tregs(\tat,\rad) = \kbrace{\atom = \tat+ \zz : \nor[\zz]\leq \rad}.
\end{align}
Therefore, we have 
\begin{align}
\max_{\atom\in\tregs(\tat,\rad)}\scap[\atom]{\cen}
&= \scap[\tat]{\cen} + \max_{\nor[\zz]\leq \rad} \scap[\zz]{\cen} \nonumber\\
&= \scap[\tat]{\cen} +  \rad\, \nor[\cen],  \label{eq:supA1}
\end{align}
where the last equality is a consequence of the tightness of the Cauchy-Schwarz inequality. The joint screening test \eqref{eq:spheretest} then straightforwardly follows from \eqref{eq:supA1}.


\subsection{Dome joint test \eqref{eq:dometest1}-\eqref{eq:dometest2}}

We assume $\kvvbar{\tat}_2=1$. We first note that the dome region $\tregd(\tat,\spcmin)$ can be written as
\begin{align}
\tregd(\tat,\spcmin) = \kbrace{\spc \tat+ \zz :  
\zz\in\Zc_\spc, 
\spcmin\leq \spc\leq 1}, \label{eq:autredefdome}
\end{align}
where 
\begin{align}
\Zc_\spc = \kbrace{\zz : \zz\in(\spa[\tat])^\perp,\nor[\zz]\leq\sqrt{1-\spc^2}}. \label{eq:defZalpha}
\end{align}
Therefore, we have 
\begin{align}
\max_{\atom\in\tregd(\tat,\spcmin)}\kparen{\scap[\atom]{\cen}}
&=  \max_{\spcmin\leq \spc\leq 1}\kparen{ \spc\  \scap[\tat]{\cen} + \max_{\zz\in\Zc_\spc} \kparen{\scap[\zz]{\cen}}}\nonumber\\
&=  \max_{\spcmin\leq \spc\leq 1} \kparen{\spc\  \scap[\tat]{\cen} + \max_{\zz\in\Zc_\spc}  \kparen{\scap[\z]{P_{\tat}^\perp(\cen)}}}\nonumber\\
&=  \underbrace{\max_{\spcmin\leq \spc\leq 1} \kparen{\spc\  \scap[\tat]{\cen} + \sqrt{1-\spc^2}\, \nor[P_{\tat}^\perp(\cen)]}}_{\triangleq g(\spcmin)}, \nonumber
\end{align}
where $P_{\tat}^\perp(\cdot)$ denotes the projector onto $(\spa[\tat])^\perp$. The last equality is a consequence of the tightness of the Cauchy-Schwarz inequality. 

The joint dome test can thus simply be rewritten as
\begin{align}
g(\spcmin)<\radss. \label{eq:lemma_jointtest_general}
\end{align}
Using Lemma \ref{lemma:propfunctions} with $A=\frac{\scap[\tat]{\cen}}{\nor[\cen]}$,\footnote{We use the fact that $\nor[P_{\tat}^\perp(\cen)]=\sqrt{\nor[\cen]^2-\scap[\tat]{\cen}^2}$.} we obtain that 
\begin{align}
g(\spcmin)&=
\left\{
\begin{array}{ll}
\nor[\cen] & \mbox{if $A<\frac{\scap[\tat]{\cen}}{\nor[\cen]}$}\\
\delta \scap[\tat]{\cen} + \sqrt{1-\spcmin^2}\sqrt{\nor[\cen]^2-\scap[\tat]{\cen}^2} & \mbox{otherwise}
\end{array}
\right. .\label{eq:expression_g}
\end{align}
The joint group test \eqref{eq:dometest1}-\eqref{eq:dometest2} can then be shown as follows. 
First, satisfying \eqref{eq:lemma_jointtest_general} necessarily requires that 
\begin{align}
\tau > \min_{\tilde{\delta}\in[-1,1]} g(\tilde{\delta}) = g(1) = \scap[\tat]{\cen}. 
\end{align}
This corresponds to the condition enforced by \eqref{eq:dometest1}. 

Moreover, we have from Lemma~\ref{lemme:ratiotaunorcen} that 
\begin{align}
\tau\leq \nor[\cen]\nonumber
\end{align}
provided that $\tau$ is associated to the radius of a safe sphere. Therefore, if \eqref{eq:dometest1} holds, owing to the continuity of $g$ and the fact that it is strictly decreasing over $[\frac{\scap[\tat]{\cen}}{\nor[\cen]},1]$ (see Lemma~\ref{lemma:propfunctions}), there exists $\spcmin_{\tat,\cen}\in [\frac{\scap[\tat]{\cen}}{\nor[\cen]},1]$ such that $g(\spcmin_{\tat,\cen}) = \tau. 
$
 Using the expression of $g$ over $[\frac{\scap[\tat]{\cen}}{\nor[\cen]},1]$ in \eqref{eq:expression_g}, we find 
\begin{align}
\spcmin_{\tat,\cen} &= \frac{\scap[\tat]{\cen} \radss + \sqrt{\nor[\cen]^2-\scap[\tat]{\cen}^2} \sqrt{\nor[\cen]^2-\radss^2}}{\nor[\cen]^2}. \label{eq:lemma_expression_spcmi_tc}
\end{align}
Invoking the strict decrease of $g$ over $[\frac{\scap[\tat]{\cen}}{\nor[\cen]},1]$ (see Lemma~\ref{lemma:propfunctions}), we have that $\tau = g(\spcmin_{\tat,\cen})>g(\spcmin)$ if and only if 
\begin{align}
\spcmin>\spcmin_{\tat,\cen}. \nonumber
\end{align}
Combining this condition with the expression of $\spcmin_{\tat,\cen}$ in \eqref{eq:lemma_expression_spcmi_tc}, we obtain \eqref{eq:dometest2}.

\section{Proof of Lemma 1}\label{app:proof_relation_mindome_minsphere}

On the one hand, the dome of smallest volume including all the elements of $\targetset$ is given by $\tregd(\tat^\star,\spcmin^\star)$ with
\begin{align}
\tat^\star &= \kargmax_{\tat: \nor[\tat]=1} \min_{\atom \in\targetset} \scap[\tat]{\atom},\label{eq:mindome1}\\
\spcmin^\star &= \min_{\atom \in\targetset} \scap[\tat^\star]{\atom}.\label{eq:mindome2}
\end{align}
On the other hand, 
 the minimum-volume sphere covering all the elements of $\targetset$ is given by $\tregs(\tatt^\star,\rad^\star)$ with
\begin{align}
\tatt^\star &= \kargmin_{\tatt} \max_{\atom\in\targetset} \nor[\atom-\tatt],\label{eq:problem1}\\
\rad^\star&= \max_{\atom\in\targetset} \nor[\atom-\tatt^\star]. \label{eq:problem2}
\end{align}
We first show that the optimal parameters $(\tat^\star,\spcmin^\star)$ and $(\tatt^\star, \rad^\star)$ are related as follows
\begin{align}
\tatt^\star &= \bar{\spcmin}\, \tat^\star, \label{eq:opttatsphere}\\
\rad^\star &= \sqrt{1-\bar{\spcmin}^2},\label{eq:optradsphere}
\end{align}
where
\begin{align}
\bar{\spcmin} = \max(0,\spcmin^\star). \label{eq:defspcminbar}
\end{align}
Indeed, setting $\tatt = \beta \optimvar$ with $\beta\geq 0$ and $\nor[\optimvar]=1$, \eqref{eq:problem1} can also be rewritten as 
\begin{align}
(\beta^\star, \optimvar^\star) 
&= \kargmin_{\beta\geq 0,\nor[\optimvar]=1} \max_{\atom\in\targetset} \nor[\atom-\beta\optimvar]^2,\nonumber\\
&= \kargmin_{\beta\geq 0,\nor[\optimvar]=1} \kparen{\beta^2 - 2 \beta \min_{\atom\in\targetset} \scap[\optimvar]{\atom}}. \label{eq:problem1b}
\end{align}
From \eqref{eq:problem1b}, we clearly have that
\begin{align}
\optimvar^\star &= \kargmax_{\nor[\optimvar]=1} \min_{\atom\in\targetset} \scap[\optimvar]{\atom}. \nonumber
\end{align}
In view of \eqref{eq:mindome1}, we thus have $\optimvar^\star = \tat^\star$. Taking this fact into account, we deduce
\begin{align}
\beta^\star 
&= \kargmin_{\beta\geq 0} \kparen{\beta^2 - 2 \beta \spcmin^\star}
=  \bar{\spcmin}, \nonumber
\end{align}
where $\bar{\spcmin}$ is defined in \eqref{eq:defspcminbar}, and thus $\tatt^\star = \bar{\spcmin}\, \tat^\star$.  Plugging $\tatt^\star = \bar{\spcmin}\, \tat^\star$ in \eqref{eq:problem2} and using the definition of $\spcmin^\star$ in \eqref{eq:mindome2}, we find
\begin{align}
\rad^\star = \sqrt{1-\bar{\spcmin}^2}. \nonumber
\end{align}
This shows \eqref{eq:opttatsphere}-\eqref{eq:optradsphere}. 

We now prove the result of the lemma, that is
\begin{align}
\tregd(\tat^\star,\spcmin^\star)\subseteq \tregs(\tatt^\star,\rad^\star). \nonumber
\end{align}
This statement can equivalently be rewritten as
\begin{align}
\forall \atom\in \tregd(\tat^\star,\spcmin^\star):\, \nor[\atom-\bar{\spcmin} \tat^\star]^2 \leq 1-\bar{\spcmin}^2.  \nonumber
\end{align}
If $\bar{\spcmin}=0$, the inequality is true since $\nor[\atom]\leq$ $\forall\,\atom\in\tregd(\tat^\star,\spcmin^\star)$. 
If $\bar{\spcmin}=\spcmin^\star$, the inequality is also verified because
\begin{align}
\nor[\atom-\spcmin^\star \tat^\star]^2 
&= 1+ (\spcmin^\star)^2 - 2 \spcmin^\star \scap[\tat^\star]{\atom} \nonumber\\
&\leq 1-(\spcmin^\star)^2 \nonumber
\end{align}
where the last inequality follows from the fact that
\begin{align}
\forall \atom\in \tregd(\tat^\star,\spcmin^\star):\ \scap[\tat^\star]{\atom}\geq \spcmin^\star. \nonumber
\end{align}

\section{Miscellaneous Technical Lemmas}\label{app:technical_lemmas}

In this section, we state and prove two technical lemmas which are useful for the derivation of the dome sphere test in Appendix \ref{app:defsup}.~The first lemma (Lemma~\ref{lemma:propfunctions}) characterizes the properties of a particular function.~The second lemma (Lemma~\ref{lemme:ratiotaunorcen}) establishes a relationship between the radius parameter $\tau$ and the center $\cen$ of a safe sphere. 

\begin{lemma}\label{lemma:propfunctions}
If $A\in[-1,1]$ (resp. $A\in(-1,1)$), the function
\begin{align}
f(\xi)  &= A\, \xi + \sqrt{1-A^2} \sqrt{1-\xi^2} \label{app:deffxi}
\end{align}
is concave (resp. strictly concave) over the interval $[-1,1]$. Moreover, the function $g(\xi) 
\triangleq \max_{\xi \leq \xi'\leq 1} f(\xi')$ can be written as follows in the interval $\xi\in[-1,1]$:
\begin{align}
g(\xi) 
&= \left\{
\begin{array}{ll}
1  & \mbox{if $\xi< A$}\\
A\, \xi + \sqrt{1-A^2} \sqrt{1-\xi^2} & \mbox{otherwise}
\end{array} 
\right. . \label{app:defgxi}
\end{align}
$g(\xi)$ is concave and non-increasing over the interval $[-1,1]$. If $A\in(-1,1)$, it is strictly concave and strictly decreasing over the interval $[A, 1]$.
\end{lemma}

\textit{Proof}: The strict concavity of $f(\xi)$ over $[-1,1]$ for $A\in(-1,1)$ can be proved by showing that its second derivative is strictly negative over this interval. Now, we have
\begin{align}
f''(\xi)
&= -\sqrt{\frac{1-A^2}{(1-\xi^2)^3}}<0\quad \forall\, \xi\in[-1,1]\nonumber.
\end{align}
The concavity of $f(\xi)$ over $[-1,1]$ for $A\in[-1,1]$ then follows by noticing that $f(\xi)$ is a linear function when $A=\pm1$.

The expression of $g(\xi)$ in \eqref{app:defgxi} can be found as follows. If $A=1$, we have 
\begin{align}
\max_{\xi \leq \xi'\leq 1} f(\xi') &= \max_{\xi \leq \xi'\leq 1} \xi' = 1 \quad \forall\xi\in[-1,1]. 
\end{align}
This corresponds to the definition of $g(\xi)$ in \eqref{app:defgxi}. If $A=-1$, we have 
\begin{align}
\max_{\xi \leq \xi'\leq 1} f(\xi') &= \max_{\xi \leq \xi'\leq 1} -\xi' = -\xi \quad \forall\xi\in[-1,1]. 
\end{align}
This again corresponds to the definition of $g(\xi)$ in \eqref{app:defgxi}. Let us finally consider the case where $A\in(-1,1)$. Using the first part of the lemma, we have that $f(\xi')$ is strictly concave over $\xi'\in[-1,1]$. Simple calculus shows that $\xi'=A$ cancels out the first derivative of $f(\xi')$. We thus have
\begin{align}
1 = f(A) = \max_{-1\leq \xi'\leq 1} f(\xi').
\end{align}
Using standard optimality conditions for concave problems, we then obtain \eqref{app:defgxi}. 

Finally, the concavity and the non-increasing (resp. the strict concavity and the strictly decreasing) nature of $g(\xi)$ over $[-1,1]$ (resp. $[A,1]$) for $A\in[-1,1]$ (resp. $A\in(-1,1)$) follows from its definition of $g(\xi)$ and the  concavity (resp. strict concavity) of $f(\xi)$. 

\hfill$\square$

\begin{lemma}\label{lemme:ratiotaunorcen}
Let $\sreg=\kbrace{\thetab: \nor[\thetab-\cen]\leq 1-\radss}$ be a safe sphere for problem \eqref{eq:Dual_Problem} with $\lambda<\lambda_{\mathrm{max}}$ (that is $\thetab^\star_\lambda\in \sreg$). Then, we have
\begin{align}
\radss \leq \nor[\cen]. \label{eq:mainres_lemtaunorc}
\end{align}
\end{lemma}
\textit{Proof}: If $\sreg=\kbrace{\thetab: \nor[\thetab-\cen]\leq 1-\radss}$ is a safe region, then 
\begin{align}
\nor[\thetab^\star_\lambda - \cen]\leq 1-\radss, 
\end{align}
which leads, by using a triangle inequality, to
\begin{align}
\nor[\thetab^\star_\lambda] -1+\radss \leq \nor[\cen].  \label{eq:lemmtaunorc_intr1}
\end{align}
Now, if $\lambda<\lambda_{\mathrm{max}}$, we have
\begin{align}
\nor[\thetab^\star_\lambda]\geq 1. \label{eq:northeta>1}
\end{align}
The latter claim follows from the following arguments: if $\lambda<\lambda_{\mathrm{max}}$, we necessarily have $\x^\star_\lambda(i)>0$ for some $i\in[1\ldots\dimx]$. 
%
From the optimality condition \eqref{eq:primal feasibility 2}, we have for such $i$: $\scap[\atom_i]{\thetab^\star_\lambda}=1$. Hence, we obtain \eqref{eq:northeta>1} by using the Cauchy-Schwarz inequality:
\begin{align}
\nor[\thetab^\star_\lambda] \geq \scap[\atom_i]{\thetab^\star_\lambda}=1. \nonumber
\end{align}
Finally, we obtain the main result \eqref{eq:mainres_lemtaunorc} by combining \eqref{eq:lemmtaunorc_intr1} and \eqref{eq:northeta>1}. 
$\hfill \square$\\[0.2cm]

\bibliographystyle{IEEEbib}
\bibliography{MyBibli}


\end{document}